%% file: felsner.tex
\documentclass[journal]{IEEEtran}

\usepackage[pdftex]{graphicx}
\usepackage{tikz,pgfplots}
\pgfplotsset{compat=1.15}
\usepackage{subfig}

\usepackage{amsmath}
\renewcommand*\vec[1]{\mathbf{#1}}
\usepackage{bbold}
\usepackage{mathabx}

\usepackage{color}

\usepackage[english]{babel}

\usepackage[utf8]{inputenc} 
\DeclareUnicodeCharacter{200B}{{\hskip 0pt}}

\newcommand{\etal}{\textit{et~al.}}

\title{Reconstruction of Voxels with Position- and Angle-Dependent Weightings}

\author{Lina~Felsner, 
	Tobias~W\"urfl, Christopher~Syben, Philipp~Roser, Alexander~Preuhs, 
	Andreas~Maier, Christian~Riess
	\thanks{All authors are with the Pattern Recognition Lab, University of Erlangen-Nuremberg, Germany.\newline\{lina.felsner, christian.riess\}@fau.de}
	\thanks{L.~Felsner, A.~Maier are with the International Max Planck Research School - Physics of Light}
}

\usepackage{fancyhdr}
\fancypagestyle{title}{%
	\setlength{\headheight}{22pt}%
	\fancyhf{}
	\fancyhead[L]{\small\color{red}\begin{tabular}{ll}This paper is a corrected version of the original work​\\that was published at the CT meeting 2020\end{tabular}}
	\fancyhead[R]{\scriptsize\thepage}
}%

\begin{document}

\maketitle
\thispagestyle{title}

\begin{abstract}
The reconstruction problem of voxels with individual weightings can be modeled a position- and angle- dependent function in the forward-projection.
This changes the system matrix and 
prohibits to use standard filtered backprojection.

In this work we first formulate this reconstruction problem 
in terms of a system matrix and weighting part.
We compute the pseudoinverse and show that the solution is rank-deficient and
hence very ill posed. This is a fundamental limitation for reconstruction.
We then derive an iterative solution and experimentally show its superiority
to any closed-form solution.

\end{abstract}

\IEEEpeerreviewmaketitle

\section{Introduction}
\IEEEPARstart{T}{he}
theory of X-ray computed tomography is well established~\cite{natterer1986computerized,zeng2010medical}.
The reconstruction problem is oftentimes expressed in matrix notation as a linear system of equations
\begin{equation}
\vec{p} = \mathbf{A}\vec{x} \enspace,
\label{eq:p=Ax}
\end{equation}
where $\vec{p}$ are the projections, $\vec{x}$ the unknown voxel content, and $\mathbf{A}$ the system matrix.
Each element $a_{i,j}$ of the system matrix $\mathbf{A}$ describes the contribution of a particular voxel to a particular projection.
$\mathbf{A}$ is typically full-rank, and tall (i.e., it has more rows
than columns), which allows to solve Eqn.~\ref{eq:p=Ax} via the left-hand sided pseudoinverse $\mathbf{A}^\ddagger$
of $\mathbf{A}$ (cf. Tab.~\ref{tab:moore_penrose}). This leads to the solution
\begin{equation}
\vec{x} = \left(\mathbf{A}^\top\mathbf{A}\right)^{-1} \mathbf{A}^\top\vec{p} \enspace,
\label{eq:A+}
\end{equation}
which is commonly known as backprojection filtering where
$\left(\mathbf{A}^\top\mathbf{A}\right)^{-1}$ represents the filter and $\mathbf{A}^\top$ the backprojection.\\

One basic assumption of the reconstruction problem is that the unknown quantity $\vec{x}$ is constant under all projections.
However, there exist setups with more complicated signal formation, such that
this assumption does not hold.  One example is the sensitivity of the phase
shift in a Talbot-Lau grating
interferometer~\cite{engelhardt2007high,donath2009inverse}. The measured
diffraction is a linear ramp that depends on the distance between the voxel and
two gratings that bound the measurement space.  Hence, parts of a large object
that are closer to the source contribute a much stronger signal than parts that
are closer to the detector. This ramp has to be considered as a
position-dependent function in single
projections~\cite{engelhardt2007high,donath2009inverse} and as a position- and
angle-dependent function in tomography~\cite{chabior2012grating}.
Another example is the X-ray dark-field signal~\cite{Jesen10:DXD,Revol12:OSX}.
Here, each voxel is measured an angle-dependent linear combination of isotropic
and anisotropic scattering, which can also be represented as an angle-dependent
weighting in the reconstruction~\cite{bayer2014reconstruction}.

Such voxel characteristics can also be formulated as a component of the system
itself.  Since the system matrix already describes a position- and
angle-dependent mapping, the integration of such specific weights in the
forward-projector via multiplication is straightforward.

However, the modification of the system matrix prohibits the use of standard
reconstruction algorithms, since the filter-kernel and adjoint operator are not known.
Therefore, we propose to decompose the weighted system matrix into a pure
system matrix and a weighting matrix.  This allows use the known filter
and backprojection with additional operations defined by the weights. 

In this paper, we formulate this reconstruction problem and show  in
Sec.~\ref{sec:analytic_reco} why the solution to the decomposed system is
ill-conditioned. We carry out a simple experiment to explicitly calculate the different
possible pseudoinverses. This shows that an exact solution cannot be obtained
unless the weighted system matrix is small enough to directly calculate the
pseudoinverse.
In a second step, we derive in Sec.~\ref{sec:algebraic_reco} the iterative 
update formula and show that the solution can be iteratively approximated.
We report experiments to compare the iterative scheme and its L1-weighted
variant to backprojection filtering and an approximate solution.
Finally, we conclude our findings and give a brief outlook in Sec.~\ref{sec:conclusion}.

\section{Calculation of the Pseudoinverse: \\Origin and Impact of its Ill-conditioning}\label{sec:analytic_reco}
We investigate the inclusion of a position- and angle-dependent function into the reconstruction problem.
Inclusion of the weights in the forward-projector 
can be formulated as  
\begin{equation}
p(t, \theta) = \int f(x, y) \cdot g(x, y, \theta) \; \textup{d} \vec{r} \enspace ,
\label{eq:proj_sens}
\end{equation}
where $p(t, \theta)$ is the projection at detector position $t$ under rotation $\theta$ along the projection ray $\vec{r}$.
The object is given as function $f(x,y)$. The function $g(x,y, \theta)$ is a position- and angle-dependent multiplicative factor. 
In matrix notation, the equivalent formulation is
\begin{equation}
\vec{p} = \tilde{\mathbf{A}}\vec{x} \enspace,
\end{equation}
where $\tilde{\mathbf{A}}$ denotes the system matrix including the weights given by $g(x,y, \theta)$.
According to Eqn.~\ref{eq:A+}, the solution can be calculated with the pseudoinverse $\tilde{\mathbf{A}}^+$.
The standard system matrix $\mathbf{A}$ allows to derive the adjoint operator and
filter function with the Fourier-Slice theorem. However, an analogous
interpretation for
$\left(\tilde{\mathbf{A}}^\top\tilde{\mathbf{A}}\right)^{-1}$ and
$\tilde{\mathbf{A}}^\top$ is not available and hence unknown.

Nevertheless, the weighted reconstruction problem can be analytically examined
by deconstructing $\tilde{\mathbf{A}} = \mathbf{B}\mathbf{W}$ into a pure system matrix $\mathbf{B}$ and a weight
matrix $\mathbf{W}$. The weights $\mathbf{W}$ shall be individually chosen per
voxel $x_i$ and per rotation angle $\theta_j$. Hence, 
$\mathbf{W}$ consists for each angle $\theta_j$ of a diagonal submatrix
$\mathbf{W}_{\theta_j} = \text{diag}\left(w_{j,1}, w_{j,2}, \dots , w_{j,V}\right)$.
The full matrix $\mathbf{W}$ is given on the right side of Eqn.~\ref{eq:AandW}. The
angle-dependency in $\mathbf{W}$ requires to construct the system matrix
$\mathbf{B}$ slightly different from traditional system matrices $\mathbf{A}$:
$\mathbf{B}$ is a block-diagonal matrix, where each block contains the
traditional system matrix under one specific angle $\theta_j$. Hence,
\begin{eqnarray}
\mathbf{B} = 
\begin{bmatrix} 
\mathbf{A}_{\theta_1}  & &  & \\ 
  & \mathbf{A}_{\theta_2}&  & \\
  & & \ddots & \\
  & &  &  \mathbf{A}_{\theta_\Omega}
\end{bmatrix}
& 
\mathbf{W} = 
\begin{bmatrix} 
\mathbf{W}_{\theta_1} \\ 
\mathbf{W}_{\theta_2} \\ 
\vdots \\
\mathbf{W}_{\theta_\Omega}
\end{bmatrix}\enspace,
\label{eq:AandW}
\end{eqnarray}
and $\tilde{\mathbf{A}} = \mathbf{B} \mathbf{W}$ as stated above. 
$\mathbf{W}$ has full rank due to its design from diagonal matrices.
$\mathbf{B}$ has full rank due to its block-diagonal structure and because
every row within one block models an individual ray with individual voxel
contributions. 
The full reconstruction equation is
\begin{equation}
\vec{p} = \mathbf{B}\mathbf{W}\vec{x} \enspace .
\label{eq:p=BWx}
\end{equation}

As in traditional reconstruction, solving Eqn.~\ref{eq:p=BWx} for $\vec{x}$
requires to multiply the pseudoinverse from the left. To this end, two
approaches could be considered, namely either
\begin{equation}
\mathbf{W}^+\mathbf{B}^+ \mathbf{p} = \mathbf{W}^+\mathbf{B}^+ \mathbf{B}\mathbf{W} \mathbf{x} 
\label{eqn:pseudo_inv_naiv}
\end{equation}
or 
\begin{equation}
(\mathbf{B}\mathbf{W})^+ \mathbf{p} = (\mathbf{B}\mathbf{W})^+ (\mathbf{B}\mathbf{W}) \mathbf{x}\enspace.
\label{eqn:generaliz_pseudo_inv}
\end{equation}

However, it is important to realize that for the dimensions of $\mathbf{B}$ and
$\mathbf{W}$, the individual inversion of these matrices in
Eqn.~\ref{eqn:pseudo_inv_naiv} does not
isolate $\mathbf{x}$, and can hence not be used. This
follows from the fact that $\mathbf{B}$ is a broad matrix with dimensions
$N\Omega \times V\Omega$ with $N < V$, since there are by definition more
voxels $V$ than detector pixels $N$. In this case, the right-hand sided Moore-Penrose pseudoinverse $\mathbf{B}^\dagger$ has to be used (see Tab.~\ref{tab:moore_penrose}).
However, left-multiplying the right-hand side pseudoinverse $\mathbf{B}^\dagger$ leads 
to the term
$\mathbf{B}^\top\left(\mathbf{B}\mathbf{B}^\top\right)^{-1}\mathbf{B} \ne
\mathbf{I}$ and hence does not isolate $\mathbf{x}$ as intended. 
As a consequence, it is necessary to calculate the general pseudoinverse from
Eqn.~\ref{eqn:generaliz_pseudo_inv}. Note that this
insight also implies that
\begin{equation}
(\mathbf{B}\mathbf{W})^+ \neq \mathbf{W}^{+}\mathbf{B}^+ \enspace.
\end{equation}

\textit{Lemma.}
	The generalized pseudoinverse of the matrices $\mathbf{B}$ and $\mathbf{W}$ 
\label{lemma}
\begin{align}
(\mathbf{B}\mathbf{W})^+ 
&= \left(\mathbf{B}^{+}\mathbf{B}\mathbf{W}\right)^{+}\left(\mathbf{B}\mathbf{W}\mathbf{W}^{+}\right)^{+}
\enspace .
\end{align}
does not have full rank. Thus, we have more degrees of freedom than linearly
independent equations, which makes the system of equations ill-conditioned.
On top we also have $\mathbf{B}^+ \mathbf{B}$, which does not isolate $\vec{x}$. Thus the term is even worse than the other pseudoinverse ($\mathbf{W}^+\mathbf{B}^+ \mathbf{p}$).
	
\textit{Proof.}
We show that the rows in $(\mathbf{B}\mathbf{W})^+$ are linearly dependent, from which
the remaining statements follow. The pseudoinverse of the matrix product
$(\mathbf{B}\mathbf{W})^+$ is defined as~\cite{matrixcookbook}
\begin{align}
(\mathbf{B}\mathbf{W})^+ 
=& \left(\mathbf{B}^+\mathbf{B}\mathbf{W}\right)^{+}\left(\mathbf{B}\mathbf{W}\mathbf{W}^+\right)^{+} \label{eq:pseudo_two}\\
=& \left(\mathbf{B}^\dagger\mathbf{B}\mathbf{W}\right)^{+}\left(\mathbf{B}\mathbf{W}\mathbf{W}^\ddagger\right)^{+} \label{eq:pseudo_twoTwo}\\
=&\!\left(\mathbf{B}^\top \left(\mathbf{B}\mathbf{B}^\top\right)^{-1}\mathbf{B}\mathbf{W}\right)^{+} \nonumber \\
&\enspace \cdot \!\left(\mathbf{B}\mathbf{W}\left(\mathbf{W}^\top\mathbf{W}\right)^{-1}\mathbf{W}^\top\right)^{+}\!,\label{eq:pseudoI}
\end{align}
where we inserted between Eqn.~\ref{eq:pseudo_two} to \ref{eq:pseudoI} the
respective left- and right-hand sided Moore-Penrose pseudoinverses.

\begin{table}[tp]
	\centering
	\caption{List of symbols} 
	\label{tab:symbols}
	\begin{tabular}{llp{5.2cm}}
		\textbf{Symbol} & \textbf{Dimension} 		& \textbf{Interpretation}\\
		\hline
		\hline
		$V$ \phantom{$x^{2^2}$}			& 							& Number of voxels\\
		$\Omega$ 		& 							& Number of projection angles\\
		$\theta_i$		&							& One projection angle $\left(\forall i \in \{1, ..., \Omega\}\right)$\\
		$N$ 			& 							& Number of detector pixels\\
		\hline
		$\vec{x}$ \phantom{$x^{2^2}$}			& $ \mathbb{R}^{V \times 1}$ 		& Vector representation of the object\\
		$\vec{p}$		& $ \mathbb{R}^{N\Omega \times 1}$ 	& Vector containing the projections\\
		$\mathbf{A}$ 	& $ \mathbb{R}^{N\Omega \times V}$ 	& System matrix\\
		$\tilde{\mathbf{A}}$ & $ \mathbb{R}^{N\Omega \times V}$ & System matrix with weights\\
		$\mathbf{B}$ 	& $ \mathbb{R}^{N\Omega \times V\Omega}$& Block matrix where each block contains the system matrix for one angle\\
		$\mathbf{W}$ 	& $ \mathbb{R}^{V\Omega \times V}$	& Weight matrix \\
		\hline
		$\cdot_{i}$	\phantom{$x^{2^2}$}		&							& Element $i$ of vector $\cdot$\\
		$\cdot_{i,j}$		&						& Element $i, j$ of matrix $\cdot$\\
		$\cdot^+$ 		& 							& Pseudoinverse\\
		\hline
	\end{tabular}
\end{table}

\begin{table}[tp]
	\centering
	\caption{Moore-Penrose pseudoinverses for a matrix $\mathbf{A}$ with full rank~\cite{matrixcookbook}} 
	\label{tab:moore_penrose}
	\begin{tabular}{llll}
		\textbf{Dimension} & \textbf{Name} 		& \textbf{Rank} & \textbf{Pseudoinverse} \\
		\hline
		\hline
		$n \times n$ & 	Square	& $n$\phantom{$x^{2^2}$}				& $\mathbf{A}^+ = \mathbf{A}^{-1}$ \\
		$n \times m$ & 	Broad	& $n$			& $\mathbf{A}^+ = \mathbf{A}^\dagger = \mathbf{A}^\top\left(\mathbf{A}\mathbf{A}^\top\right)^{-1}$\\
		$n \times m$ & 	Tall	& $m$			& $\mathbf{A}^+ = \mathbf{A}^\ddagger = \left(\mathbf{A}^\top\mathbf{A}\right)^{-1}\mathbf{A}^\top$\\
		\hline
	\end{tabular}
\end{table}

The linear dependency occurs in the product $\mathbf{WW}^\ddagger =
\mathbf{W}\left(\mathbf{W}^\top\mathbf{W}\right)^{-1}\mathbf{W}^\top$. In order
to see this, we first note that $\left(\mathbf{W}^\top\mathbf{W}\right)^{-1}$ is a diagonal
matrix due to the diagonal structure of the submatrices of $\mathbf{W}$.
\begin{align}
&\left(\mathbf{W}^\top\mathbf{W}\right)^{-1} = \nonumber\\
&
\left(\begin{bmatrix} 
\text{diag}\left(w_{1,1},
\ldots , w_{1,V}\right) \\ 
\vdots \\
\text{diag}\left(w_{\Omega,1},
\ldots , w_{\Omega,V}\right)
\end{bmatrix}^\top \begin{bmatrix} 
\text{diag}\left(w_{1,1},
\ldots , w_{1,V} \right)\\ 
\vdots \\
\text{diag}\left(w_{\Omega,1},
\ldots , w_{\Omega,V}\right)
\end{bmatrix}\right)^{-1}\\ 
  &=
 \text{diag}\left(1/\sum_{i}w_{i,1}^2, 1/\sum_{i}w_{i,2}^2, \ldots , 1/\sum_{i}w_{i,V}^2\right) 
\end{align}
Due to the commutativity of diagonal matrices we have the equality 
$\mathbf{W} \left( \mathbf{W}^\top \mathbf{W} \right)^{-1} \mathbf{W}^\top =
\left( \mathbf{W}^\top \mathbf{W} \right)^{-1} \mathbf{W}\mathbf{W}^\top$. Here,
\begin{align}
\mathbf{W}\mathbf{W}^\top &= 
\begin{bmatrix} 
\mathbf{W}_{\theta_1}\mathbf{W}_{\theta_1} & \mathbf{W}_{\theta_1}\mathbf{W}_{\theta_2} & 
\dots & \mathbf{W}_{\theta_1}\mathbf{W}_{\theta_\Omega} \\ 
\mathbf{W}_{\theta_2}\mathbf{W}_{\theta_1} & \mathbf{W}_{\theta_2}\mathbf{W}_{\theta_2} 
& \dots & \mathbf{W}_{\theta_2}\mathbf{W}_{\theta_\Omega}\\ 
\vdots & \vdots & \ddots & \vdots\\
\mathbf{W}_{\theta_\Omega}\mathbf{W}_{\theta_1} & \mathbf{W}_{\theta_\Omega}\mathbf{W}_{\theta_2}  & 
\dots & \mathbf{W}_{\theta_\Omega}\mathbf{W}_{\theta_\Omega}
\end{bmatrix} \enspace .
\end{align}
Each block in this matrix is again a diagonal matrix
$\mathbf{W}_{\theta_i}\mathbf{W}_{\theta_j} = \text{diag}\left(w_{i,1}w_{j,1},
w_{i,2}w_{j,2}, \dots , w_{i,V}w_{j,V}\right) \forall i,j \in \{1, \dots
\Omega\}$.  The rows and columns between the block matrices are linear
dependent.  Rows $i$ and $j$ only differ by a factor $w_{\nu_i} / w_{\nu_j}$,
and the same holds for the columns.
Thus, the matrix $\mathbf{W}\mathbf{W}^\top \in \mathbb{R}^{V\cdot \Omega
\times V\cdot \Omega }$ has only rank $V$, which makes the system of equations
highly under-determined and ill-conditioned.
\hfill $\square$\\

\begin{figure}[tp]
	\centering
	\subfloat[GT $\vec{x}$\label{subfig:gt}]{\includegraphics[width=0.2\linewidth]{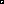}}
	\quad
	\subfloat[$\tilde{\mathbf{A}}^{+}\vec{p}$\label{subfig:1b}]{\includegraphics[width=0.2\linewidth]{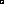}}
	\quad
	\subfloat[$\mathbf{W}^{+}\mathbf{B}^{+}\vec{p}$\label{subfig:1c}]{\includegraphics[width=0.2\linewidth]{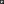}}
	\quad
	\subfloat[$\left(\mathbf{B}\mathbf{W}\right)^{+}\vec{p}$\label{subfig:1d}]{\includegraphics[width=0.2\linewidth]{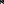}}
	\caption{Results of the pseudoinverse. 
	Fig.~a, b are windowed between $0 - 1$ [a.u.].
	Fig.~c is windowed between $0 - 0.5$ [a.u.].
	Fig.~d is windowed between $-2.5 - 12$ [a.u.].
	}
	\label{fig:toy_problem}
\end{figure}
\textbf{Experiment.}
We illustrate these findings with an example that is sufficiently small to also
calculate the pseudoinverse of the matrices directly.
We use a $4\times4$ volume, with a triangular structure of value 1 in the center as our object (Fig.~\ref{subfig:gt}). 
The system matrix uses 4 projection angels with a total of 22 projections. 
We then calculate the solutions for three different pseudoinverses.
This includes first the theoretically correct pseudoinverse of $\tilde{\mathbf{A}}$, which can be directly calculated due to the small problem size. Second, the theoretically wrong solution $\mathbf{W}^{+}\mathbf{B}^{+}\vec{p}$ , where $\mathrm{rank}(\mathbf{W}^+) = V$ and $\mathrm{rank}(\mathbf{B}^+) = N\Omega$.
Third, the theoretically wrong solution and ill-conditioned $\left(\mathbf{B}\mathbf{W}\right)^{+}\vec{p}$.

The results are shown in Fig.~\ref{fig:toy_problem}b -- \ref{fig:toy_problem}d, respectively.
The pseudoinverse of the weighted system matrix $\tilde{\mathbf{A}}$ 
delivers an almost perfect result (Fig.~\ref{subfig:1b}). The L2--distance to the ground truth $\vec{x}$ is $1.68 E-14$.
By using the theoretically incorrect pseudoinverse
$\mathbf{W}^{+}\mathbf{B}^{+}$ the overall object shape is recovered, but the
values are not correct (Fig.~\ref{subfig:1c}), with a L2--distance to the
ground truth of $1.15$.
The estimation of the theoretically wrong and ill-conditioned pseudoinverse
	completely fails to represent the object (Fig.~\ref{subfig:1d}), with a
	L2--distance to the ground truth of $16.92$.

\section{Iterative Reconstruction}\label{sec:algebraic_reco}
The first part shows that inclusion of individual weights prevents a closed-form
reconstruction. However, the original linear systems of equations in
Eqn.~\ref{eq:p=BWx} has full rank, which suggests that an iterative
solution can still be found. To this end, we derive a gradient-descent
algorithm and perform a proof-of-concept experiment where we compare the proposed method with and without additional TV regularization to 
backprojection filtering and a solution with the theoretically incorrect inversion.

Since the Moore-Penrose pseudoinverse estimates the optimal solution in a
least-squares sense, we formulate the reconstruction as an iterative
least-squares reconstruction task with the objective function
\begin{equation}
\min \frac{1}{2}||\mathbf{B}\mathbf{W}\vec{x} - \vec{p} ||^2_2 \enspace .
\label{eq:l2}
\end{equation}

Equation~\ref{eq:l2} can be iteratively solved via gradient descent.
The update-rule for the iterative algorithm is derived below following the
Kaczmarz method~\cite{kaczmarz1937}. 

One projection in the decomposed system from Eqn.~\ref{eq:p=BWx} is given as
\begin{equation}
p_i = \vec{b}_i \mathbf{W} \vec{x} \enspace,
\label{eq:line}
\end{equation}
which can be interpreted as a line in the $V$-dimensional space.
The orthogonal projection of $\vec{x}$ on this line is given as
\begin{align}
\vec{x}' &= \vec{x} + \lambda \left(\vec{b}_i \mathbf{W}\right)^\top\\
&= \vec{x} + \lambda  \mathbf{W}^\top \vec{b}_i^\top \enspace .
\label{eq:projected_point}
\end{align}
Since the projected point $\vec{x}'$ is a point on the line, Eqn.~\ref{eq:line} must also hold 
\begin{equation}
 p_i = \vec{b}_i \mathbf{W} \vec{x}' \enspace .
\label{eq:point_on_line}
\end{equation}
Inserting Eqn.~\ref{eq:projected_point} into Eqn.~\ref{eq:point_on_line} leads to
\begin{equation}
p_i = \vec{b}_i \mathbf{W} \left( \vec{x} + \lambda  \mathbf{W}^\top \vec{b}_i^\top \right)
\enspace .
\label{eqn:with_lambda}
\end{equation}
Isolating the parameter $\lambda$ in Eqn.~\ref{eqn:with_lambda} leads to
\begin{equation}
\lambda  =  \frac{\left(p_i - \vec{b}_i \mathbf{W} \vec{x}\right)}{\vec{b}_i \mathbf{W} \mathbf{W}^\top \vec{b}_i^\top} 
\enspace .
\end{equation}
This can be inserted back into Eq.~\ref{eq:point_on_line}, leading to the final update rule
\begin{equation}
\vec{x}'= \vec{x} + \frac{p_i -  \vec{b}_i \mathbf{W} \vec{x}}{\vec{b}_i \mathbf{W} \mathbf{W}^\top \vec{b}_i^\top} \mathbf{W}^\top \vec{b}_i^\top \enspace ,
\end{equation}
where $\vec{b}_i \mathbf{W} \vec{x}$ is the weighted forward-projection,
$p_i -  \vec{b}_i \mathbf{W} \vec{x}$ is the gradient,
and $\vec{b}_i \mathbf{W} \mathbf{W}^\top \vec{b}_i^\top$ is a normalization factor that contains the sum of squared weights along the ray.\\

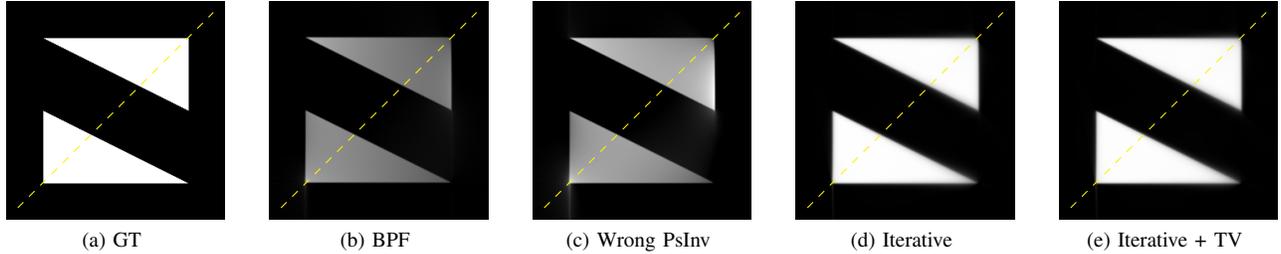
\begin{figure*}[tp]
	\centering
	\subfloat[GT \label{subfig:gt_wedge}]{
	\begin{tikzpicture}
		\node[inner sep=0pt] at (0,0){\includegraphics[width=0.16\textwidth]{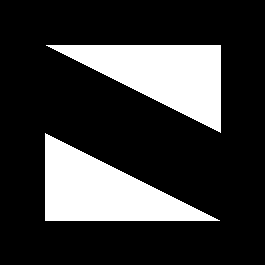} };%
		\draw[yellow, dashed] (-1.3,-1.3) -- (1.3,1.3);
	\end{tikzpicture}
	}
	\quad
	\subfloat[BPF \label{subfig:bpf}]{
	\begin{tikzpicture}
		\node[inner sep=0pt] at (0,0){\includegraphics[width=0.16\textwidth]{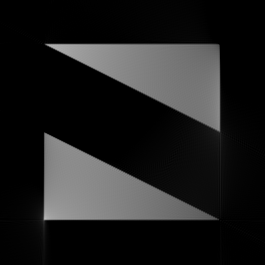} };%
		\draw[yellow, dashed] (-1.3,-1.3) -- (1.3,1.3);
	\end{tikzpicture}
	}
	\quad
	\subfloat[Wrong PsInv \label{subfig:wrong_PI}]{
	\begin{tikzpicture}
		\node[inner sep=0pt] at (0,0){\includegraphics[width=0.16\textwidth]{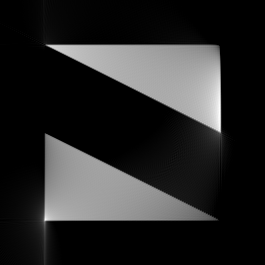} };%
		\draw[yellow, dashed] (-1.3,-1.3) -- (1.3,1.3);
	\end{tikzpicture}
	}
	\quad
	\subfloat[Iterative \label{subfig:iter}]{
	\begin{tikzpicture}
		\node[inner sep=0pt] at (0,0){\includegraphics[width=0.16\textwidth]{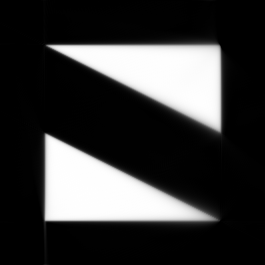} };%
		\draw[yellow, dashed] (-1.3,-1.3) -- (1.3,1.3);
	\end{tikzpicture}
	}
	\quad
	\subfloat[Iterative + TV \label{subfig:iterTV}]{
	\begin{tikzpicture}
		\node[inner sep=0pt] at (0,0){\includegraphics[width=0.16\textwidth]{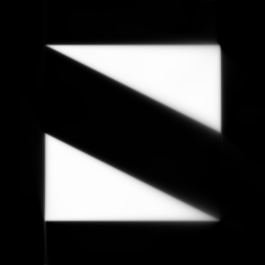} };%
		\draw[yellow, dashed] (-1.3,-1.3) -- (1.3,1.3);
	\end{tikzpicture}
	}	
	\caption{Results of the reconstructions. 
		The line plots along the yellow dashed lines are shown in Fig.~\ref{fig:wedge_line_plot}.
		Fig. c is windowed between $0 - 0.01$ [a.u.].
		All other figures are windowed between $0 - 1$ [a.u.]. }
	\label{fig:wedge}
\end{figure*}
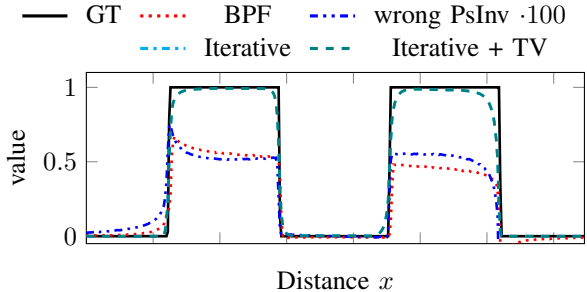
\begin{figure}[tp]
	\centering
	\vspace{-0.2cm}
	\input{line_plot.tikz} 
	\caption{Line plots along the yellow dashed lines in the reconstructions in Fig.~\ref{fig:wedge}.}
	\label{fig:wedge_line_plot}
\end{figure}

\textbf{Experiment.}
\label{sec:exp2}
In this experiment we use a more complex sample to
investigate the results of the iterative reconstruction.

The weighting function is the sensitivity ramp as it occurs in X-ray Talbot-Lau
grating
interferometers~\cite{engelhardt2007high,donath2009inverse,chabior2012grating}). 
The function is a linear ramp between a minimal and maximal sensitivity value,
which rotates along with the source and detector.  We use an experimental
setting that is similar as in Chabior~\etal~\cite{chabior2012grating}, with a
phantom that consists of two wedges in a parallel-beam geometry. We use a
$256\times 256$ voxel volume with $360$ projections over an angular range of
$\pi$.

We reconstruct the object with various approaches. First, a backprojection and
then filtering from Eqn.~\ref{eq:A+} that ignores the weights (``BPF''). Second, the
incorrect pseudoinverse $\mathbf{W}^+\mathbf{B}^+$ from Eqn.~\ref{eqn:pseudo_inv_naiv} (``Wrong PsInv'').
Third, the iterative update rule derived in this section (``Iterative''), and
fourth the iterative update rule with additional TV regularization (``Iterative
+ TV'').  The mathematical representation of the algorithms is listed in
Tab.~\ref{tab:eval}.
For the iterative reconstruction, a step size of $0.5$ is used. An adaptive
step size is used for the TV regularization. Both iterative algorithms use
$100$ iterations.

Figure~\ref{fig:wedge} shows the ground truth and the reconstruction results.
The line plots along the yellow dashed line are shown in
Fig.~\ref{fig:wedge_line_plot}.  The root mean square error (RMSE) and
structural similarity (SSIM) are reported in the right columns of
Tab.~\ref{tab:eval}.
Severe artifacts can be observed in the reconstruction with the backprojection
and filtering that ignores the weighting function. It exhibits a global
offset in the reconstructed values, and the object is not homogeneous.
The reconstruction with the incorrect pseudoinverse shows similar artifacts. 
Also here, the scaling of the reconstructed values is only half of the actual
values. The RMSE is twice as large as for backprojection and filtering, and the
SSIM is decreased by several orders of magnitude.
The iterative reconstruction is visually and quantitatively very good. 
The TV regularization changes the quantitative result only at the fifth decimal position.
Visually, both reconstructions are identical with a high similarity to the ground truth. 

This only very small performance difference between iterative reconstruction
and its regularized version can be attributed to two reasons. The experiment is
neither under-determined, nor is any noise included in this experiment.  We
expect a larger improvement of TV over the pure iterative algorithm in the
presence of this cases.

\section{Conclusion and Outlook}\label{sec:conclusion}
We investigated the inclusion of a per-voxel position- and angle-dependent
weighting function in tomographic reconstruction.  First, we describe the
problem and decompose it into a system matrix and weighting matrix.
Calculating the pseudoinverse shows in a rank analysis that the solution is
very ill-conditioned. In a second part, we derive an iterative solution and
experimentally show its superiority to any possible algebraic solution.  We
conclude that it is not possible to obtain an analytic solution for the
reconstruction of voxels with general position- and angle-dependent weightings,
while experimental results for iterative reconstruction well approximate the
ground truth solution.

In future work, it will be interesting to investigate special cases of weight
functions, and in particular whether for these cases filters can be derived
analytically or learned with a Neural-Network~\cite{syben2018deriving}.

\begin{table}[tp]
	\centering
	\caption{RMSE and SSIM of the results in Fig.~\ref{fig:wedge}} 
	\label{tab:eval}
	\begin{tabular}{llll}
		\textbf{Method} &  \textbf{Equation } &\textbf{RMSE} 		& \textbf{SSIM} \\
		\hline
		\hline
		BPF \phantom{$x^{2^2}$}	 & $\mathbf{A}^\ddagger\vec{p}$ & 	$0.24$	& $0.61$		\\
		Wrong PsInv & $\mathbf{W}^\ddagger\mathbf{B}^\dagger\vec{p}$ & 	 $0.47$	& 	$1.26 E-4$		\\
		Iterative & $\min \frac{1}{2}||\mathbf{B}\mathbf{W}\vec{x} - \vec{p} ||^2_2$ 	& 	$0.06$	& $0.99$		\\
		Iterative + TV  & $\min \frac{1}{2}||\mathbf{B}\mathbf{W}\vec{x} - \vec{p} ||^2_2 + \lambda \, \text{TV} \enspace .$ & 	$0.06$	& $0.99$		\\
		\hline
	\end{tabular}
\end{table}

\bibliographystyle{IEEEtran}
\bibliography{literature}

\end{document}

%% file: line_plot.tikz
    \newenvironment{customlegend}[1][]{%
	\begingroup
	\csname pgfplots@init@cleared@structures\endcsname
	\pgfplotsset{#1}%
}{%
	\csname pgfplots@createlegend\endcsname
	\endgroup
}%
\def\addlegendimage{\csname pgfplots@addlegendimage\endcsname}

\begin{tikzpicture}
\begin{customlegend}[legend columns=3,legend style={align=left,draw=none,column sep=1ex},
legend entries={GT,  BPF, wrong PsInv $\cdot 100$, \phantom{a} , Iterative, Iterative + TV}]
\addlegendimage{color=black,line width=1.2pt,solid}
\addlegendimage{color=red, line width=1.2pt, dotted}
\addlegendimage{color=blue, line width=1.2pt, dash dot dot}  
\addlegendimage{color=white, line width=1.2pt, dash dot}
\addlegendimage{color=cyan, line width=1.2pt, dash dot}
\addlegendimage{color=teal, line width=1.2pt, dashed}
\end{customlegend}
\end{tikzpicture}
\begin{tikzpicture}
 	\begin{axis}[%
width=\textwidth, 
	height=\axisdefaultheight,
	 scale = 0.4,
xmin=0,
	xmax=373,
	xlabel={Distance $x$},
	xlabel near ticks,
	xticklabels={,,},
ymin=-0.05,
	ymax=1.1,
	ylabel={value},
ylabel near ticks,
legend pos=north east
	]
	\addplot [color=black, line width=1.0pt] table[row sep=crcr]{%
0	0.000000\\1	0.000000\\2	0.000000\\3	0.000000\\4	0.000000\\5	0.000000\\6	0.000000\\7	0.000000\\8	0.000000\\9	0.000000\\10	0.000000\\11	0.000000\\12	0.000000\\13	0.000000\\14	0.000000\\15	0.000000\\16	0.000000\\17	0.000000\\18	0.000000\\19	0.000000\\20	0.000000\\21	0.000000\\22	0.000000\\23	0.000000\\24	0.000000\\25	0.000000\\26	0.000000\\27	0.000000\\28	0.000000\\29	0.000000\\30	0.000000\\31	0.000000\\32	0.000000\\33	0.000000\\34	0.000000\\35	0.000000\\36	0.000000\\37	0.000000\\38	0.000000\\39	0.000000\\40	0.000000\\41	0.000000\\42	0.000000\\43	0.000000\\44	0.000000\\45	0.000000\\46	0.000000\\47	0.000000\\48	0.000000\\49	0.000000\\50	0.000000\\51	0.000000\\52	0.000000\\53	0.000000\\54	0.000000\\55	0.000000\\56	0.000000\\57	0.000000\\58	0.000000\\59	0.000000\\60	0.000000\\61	0.030368\\62	0.777990\\63	1.000000\\64	1.000000\\65	1.000000\\66	1.000000\\67	1.000000\\68	1.000000\\69	1.000000\\70	1.000000\\71	1.000000\\72	1.000000\\73	1.000000\\74	1.000000\\75	1.000000\\76	1.000000\\77	1.000000\\78	1.000000\\79	1.000000\\80	1.000000\\81	1.000000\\82	1.000000\\83	1.000000\\84	1.000000\\85	1.000000\\86	1.000000\\87	1.000000\\88	1.000000\\89	1.000000\\90	1.000000\\91	1.000000\\92	1.000000\\93	1.000000\\94	1.000000\\95	1.000000\\96	1.000000\\97	1.000000\\98	1.000000\\99	1.000000\\100	1.000000\\101	1.000000\\102	1.000000\\103	1.000000\\104	1.000000\\105	1.000000\\106	1.000000\\107	1.000000\\108	1.000000\\109	1.000000\\110	1.000000\\111	1.000000\\112	1.000000\\113	1.000000\\114	1.000000\\115	1.000000\\116	1.000000\\117	1.000000\\118	1.000000\\119	1.000000\\120	1.000000\\121	1.000000\\122	1.000000\\123	1.000000\\124	1.000000\\125	1.000000\\126	1.000000\\127	1.000000\\128	1.000000\\129	1.000000\\130	1.000000\\131	1.000000\\132	1.000000\\133	1.000000\\134	1.000000\\135	1.000000\\136	1.000000\\137	1.000000\\138	1.000000\\139	1.000000\\140	1.000000\\141	1.000000\\142	1.000000\\143	1.000000\\144	1.000000\\145	0.138871\\146	0.000000\\147	0.000000\\148	0.000000\\149	0.000000\\150	0.000000\\151	0.000000\\152	0.000000\\153	0.000000\\154	0.000000\\155	0.000000\\156	0.000000\\157	0.000000\\158	0.000000\\159	0.000000\\160	0.000000\\161	0.000000\\162	0.000000\\163	0.000000\\164	0.000000\\165	0.000000\\166	0.000000\\167	0.000000\\168	0.000000\\169	0.000000\\170	0.000000\\171	0.000000\\172	0.000000\\173	0.000000\\174	0.000000\\175	0.000000\\176	0.000000\\177	0.000000\\178	0.000000\\179	0.000000\\180	0.000000\\181	0.000000\\182	0.000000\\183	0.000000\\184	0.000000\\185	0.000000\\186	0.000000\\187	0.000000\\188	0.000000\\189	0.000000\\190	0.000000\\191	0.000000\\192	0.000000\\193	0.000000\\194	0.000000\\195	0.000000\\196	0.000000\\197	0.000000\\198	0.000000\\199	0.000000\\200	0.000000\\201	0.000000\\202	0.000000\\203	0.000000\\204	0.000000\\205	0.000000\\206	0.000000\\207	0.000000\\208	0.000000\\209	0.000000\\210	0.000000\\211	0.000000\\212	0.000000\\213	0.000000\\214	0.000000\\215	0.000000\\216	0.000000\\217	0.000000\\218	0.000000\\219	0.000000\\220	0.000000\\221	0.000000\\222	0.000000\\223	0.000000\\224	0.000000\\225	0.000000\\226	0.000000\\227	0.442065\\228	1.000000\\229	1.000000\\230	1.000000\\231	1.000000\\232	1.000000\\233	1.000000\\234	1.000000\\235	1.000000\\236	1.000000\\237	1.000000\\238	1.000000\\239	1.000000\\240	1.000000\\241	1.000000\\242	1.000000\\243	1.000000\\244	1.000000\\245	1.000000\\246	1.000000\\247	1.000000\\248	1.000000\\249	1.000000\\250	1.000000\\251	1.000000\\252	1.000000\\253	1.000000\\254	1.000000\\255	1.000000\\256	1.000000\\257	1.000000\\258	1.000000\\259	1.000000\\260	1.000000\\261	1.000000\\262	1.000000\\263	1.000000\\264	1.000000\\265	1.000000\\266	1.000000\\267	1.000000\\268	1.000000\\269	1.000000\\270	1.000000\\271	1.000000\\272	1.000000\\273	1.000000\\274	1.000000\\275	1.000000\\276	1.000000\\277	1.000000\\278	1.000000\\279	1.000000\\280	1.000000\\281	1.000000\\282	1.000000\\283	1.000000\\284	1.000000\\285	1.000000\\286	1.000000\\287	1.000000\\288	1.000000\\289	1.000000\\290	1.000000\\291	1.000000\\292	1.000000\\293	1.000000\\294	1.000000\\295	1.000000\\296	1.000000\\297	1.000000\\298	1.000000\\299	1.000000\\300	1.000000\\301	1.000000\\302	1.000000\\303	1.000000\\304	1.000000\\305	1.000000\\306	1.000000\\307	1.000000\\308	1.000000\\309	1.000000\\310	0.347879\\311	0.000000\\312	0.000000\\313	0.000000\\314	0.000000\\315	0.000000\\316	0.000000\\317	0.000000\\318	0.000000\\319	0.000000\\320	0.000000\\321	0.000000\\322	0.000000\\323	0.000000\\324	0.000000\\325	0.000000\\326	0.000000\\327	0.000000\\328	0.000000\\329	0.000000\\330	0.000000\\331	0.000000\\332	0.000000\\333	0.000000\\334	0.000000\\335	0.000000\\336	0.000000\\337	0.000000\\338	0.000000\\339	0.000000\\340	0.000000\\341	0.000000\\342	0.000000\\343	0.000000\\344	0.000000\\345	0.000000\\346	0.000000\\347	0.000000\\348	0.000000\\349	0.000000\\350	0.000000\\351	0.000000\\352	0.000000\\353	0.000000\\354	0.000000\\355	0.000000\\356	0.000000\\357	0.000000\\358	0.000000\\359	0.000000\\360	0.000000\\361	0.000000\\362	0.000000\\363	0.000000\\364	0.000000\\365	0.000000\\366	0.000000\\367	0.000000\\368	0.000000\\369	0.000000\\370	0.000000\\371	0.000000\\372	0.000000\\373	0.000000\\ 
	};
\addplot [color=red, line width=1.0pt, dotted] table[row sep=crcr]{%
 0	0.004007\\1	0.003396\\2	0.003300\\3	0.003037\\4	0.003759\\5	0.003639\\6	0.005841\\7	0.004166\\8	0.004607\\9	0.003681\\10	0.003322\\11	0.004874\\12	0.005134\\13	0.005719\\14	0.005097\\15	0.004288\\16	0.004322\\17	0.005658\\18	0.006133\\19	0.006502\\20	0.007189\\21	0.006294\\22	0.006564\\23	0.007232\\24	0.007797\\25	0.007792\\26	0.009892\\27	0.009392\\28	0.008530\\29	0.008740\\30	0.009488\\31	0.011129\\32	0.012064\\33	0.011671\\34	0.012256\\35	0.012947\\36	0.013298\\37	0.014353\\38	0.015444\\39	0.016251\\40	0.016626\\41	0.017267\\42	0.018157\\43	0.020913\\44	0.021312\\45	0.023962\\46	0.025391\\47	0.027315\\48	0.027983\\49	0.029232\\50	0.032690\\51	0.035676\\52	0.040396\\53	0.043967\\54	0.047948\\55	0.053277\\56	0.060242\\57	0.066692\\58	0.080777\\59	0.102468\\60	0.143655\\61	0.153780\\62	0.192044\\63	0.275639\\64	0.541067\\65	0.639029\\66	0.668876\\67	0.661742\\68	0.648746\\69	0.636702\\70	0.629494\\71	0.623858\\72	0.618345\\73	0.616930\\74	0.612146\\75	0.606532\\76	0.603794\\77	0.599867\\78	0.597347\\79	0.594671\\80	0.590655\\81	0.590356\\82	0.587189\\83	0.583106\\84	0.582072\\85	0.581051\\86	0.578171\\87	0.578532\\88	0.575259\\89	0.573536\\90	0.573621\\91	0.569801\\92	0.569864\\93	0.569783\\94	0.566608\\95	0.565803\\96	0.566357\\97	0.559075\\98	0.563273\\99	0.562858\\100	0.559173\\101	0.562527\\102	0.557880\\103	0.554887\\104	0.556037\\105	0.554848\\106	0.553639\\107	0.554879\\108	0.555403\\109	0.552233\\110	0.551358\\111	0.550899\\112	0.547514\\113	0.546984\\114	0.547632\\115	0.547954\\116	0.545700\\117	0.547898\\118	0.545553\\119	0.543755\\120	0.543781\\121	0.542340\\122	0.541583\\123	0.542075\\124	0.540916\\125	0.544794\\126	0.540162\\127	0.538192\\128	0.542978\\129	0.536025\\130	0.536815\\131	0.533856\\132	0.536296\\133	0.537105\\134	0.534229\\135	0.534622\\136	0.535993\\137	0.534624\\138	0.538293\\139	0.533286\\140	0.533822\\141	0.532684\\142	0.524865\\143	0.525910\\144	0.512486\\145	0.401439\\146	0.131851\\147	0.006622\\148	-0.006440\\149	-0.015777\\150	-0.008829\\151	-0.011815\\152	-0.008601\\153	-0.011159\\154	-0.008891\\155	-0.008433\\156	-0.005994\\157	-0.008145\\158	-0.007461\\159	-0.006789\\160	-0.007455\\161	-0.007281\\162	-0.008159\\163	-0.008186\\164	-0.007993\\165	-0.006990\\166	-0.009335\\167	-0.007725\\168	-0.006889\\169	-0.006980\\170	-0.008002\\171	-0.007598\\172	-0.006429\\173	-0.006701\\174	-0.006814\\175	-0.006023\\176	-0.007401\\177	-0.006779\\178	-0.007583\\179	-0.006207\\180	-0.007346\\181	-0.006878\\182	-0.006444\\183	-0.007869\\184	-0.006895\\185	-0.006939\\186	-0.006301\\187	-0.006775\\188	-0.006625\\189	-0.006940\\190	-0.005373\\191	-0.006824\\192	-0.006567\\193	-0.006850\\194	-0.005703\\195	-0.006395\\196	-0.007363\\197	-0.006949\\198	-0.007003\\199	-0.006857\\200	-0.006301\\201	-0.006936\\202	-0.007331\\203	-0.005410\\204	-0.007450\\205	-0.007057\\206	-0.005894\\207	-0.007500\\208	-0.006669\\209	-0.006343\\210	-0.007795\\211	-0.006122\\212	-0.006583\\213	-0.007223\\214	-0.007137\\215	-0.007030\\216	-0.007333\\217	-0.007582\\218	-0.007201\\219	-0.007864\\220	-0.007036\\221	-0.007868\\222	-0.007654\\223	-0.009564\\224	-0.006094\\225	-0.006808\\226	-0.004806\\227	0.009774\\228	0.103925\\229	0.329454\\230	0.480797\\231	0.479008\\232	0.478939\\233	0.474956\\234	0.478703\\235	0.476718\\236	0.477995\\237	0.477930\\238	0.476584\\239	0.473646\\240	0.471153\\241	0.472155\\242	0.472385\\243	0.471300\\244	0.468248\\245	0.472856\\246	0.471211\\247	0.468240\\248	0.468616\\249	0.467045\\250	0.465617\\251	0.463343\\252	0.463568\\253	0.465745\\254	0.467260\\255	0.464311\\256	0.464208\\257	0.465591\\258	0.457994\\259	0.461686\\260	0.461364\\261	0.458363\\262	0.461815\\263	0.460800\\264	0.460722\\265	0.457548\\266	0.456776\\267	0.455592\\268	0.451679\\269	0.456795\\270	0.456052\\271	0.453665\\272	0.455267\\273	0.450292\\274	0.447821\\275	0.450449\\276	0.449067\\277	0.447918\\278	0.453576\\279	0.446685\\280	0.444543\\281	0.444357\\282	0.437472\\283	0.442684\\284	0.442110\\285	0.439915\\286	0.440718\\287	0.437701\\288	0.432983\\289	0.431892\\290	0.431523\\291	0.429775\\292	0.431710\\293	0.426225\\294	0.424376\\295	0.422835\\296	0.417232\\297	0.415990\\298	0.414016\\299	0.410321\\300	0.406336\\301	0.403491\\302	0.399621\\303	0.389654\\304	0.383319\\305	0.374326\\306	0.362724\\307	0.350500\\308	0.298114\\309	0.185792\\310	-0.046315\\311	-0.159429\\312	-0.157502\\313	-0.106940\\314	-0.092768\\315	-0.084375\\316	-0.072863\\317	-0.066460\\318	-0.062662\\319	-0.059525\\320	-0.052860\\321	-0.049426\\322	-0.046906\\323	-0.039844\\324	-0.039783\\325	-0.038417\\326	-0.036022\\327	-0.035231\\328	-0.033620\\329	-0.031243\\330	-0.027898\\331	-0.028052\\332	-0.028252\\333	-0.026053\\334	-0.026727\\335	-0.025559\\336	-0.023455\\337	-0.020913\\338	-0.021802\\339	-0.022345\\340	-0.020903\\341	-0.020621\\342	-0.019125\\343	-0.018561\\344	-0.017370\\345	-0.017105\\346	-0.017526\\347	-0.019438\\348	-0.016953\\349	-0.015233\\350	-0.015958\\351	-0.014483\\352	-0.014745\\353	-0.014781\\354	-0.014208\\355	-0.013874\\356	-0.012290\\357	-0.012037\\358	-0.011783\\359	-0.012286\\360	-0.014723\\361	-0.012839\\362	-0.012250\\363	-0.011407\\364	-0.012090\\365	-0.011479\\366	-0.011317\\367	-0.014304\\368	-0.009773\\369	-0.009428\\370	-0.007784\\371	-0.008301\\372	-0.009494\\373	-0.008353\\374	0.000000\\
	};	
\addplot [color=blue, line width=1.0pt, dash dot dot, y filter/.code={\pgfmathparse{\pgfmathresult*100.}\pgfmathresult}] table[row sep=crcr]{%
0	0.0001759\\1	0.0002370\\2	0.0002528\\3	0.0002415\\4	0.0002647\\5	0.0002656\\6	0.0002765\\7	0.0002324\\8	0.0002903\\9	0.0002929\\10	0.0002728\\11	0.0003029\\12	0.0003054\\13	0.0003186\\14	0.0003222\\15	0.0003289\\16	0.0003336\\17	0.0003233\\18	0.0003440\\19	0.0003543\\20	0.0003787\\21	0.0003912\\22	0.0003903\\23	0.0004089\\24	0.0004036\\25	0.0003958\\26	0.0004126\\27	0.0004210\\28	0.0004521\\29	0.0004628\\30	0.0004793\\31	0.0004893\\32	0.0004954\\33	0.0005039\\34	0.0005144\\35	0.0005489\\36	0.0005612\\37	0.0005837\\38	0.0005998\\39	0.0006152\\40	0.0006435\\41	0.0006585\\42	0.0006950\\43	0.0007207\\44	0.0007500\\45	0.0007836\\46	0.0008183\\47	0.0008664\\48	0.0009106\\49	0.0009687\\50	0.0010216\\51	0.0010772\\52	0.0011483\\53	0.0012299\\54	0.0013317\\55	0.0014505\\56	0.0016126\\57	0.0018127\\58	0.0020197\\59	0.0026004\\60	0.0036821\\61	0.0057397\\62	0.0073349\\63	0.0075111\\64	0.0070509\\65	0.0067349\\66	0.0064959\\67	0.0063241\\68	0.0061685\\69	0.0060566\\70	0.0059660\\71	0.0058929\\72	0.0058022\\73	0.0057544\\74	0.0056975\\75	0.0056136\\76	0.0055940\\77	0.0055571\\78	0.0055073\\79	0.0054821\\80	0.0054578\\81	0.0054314\\82	0.0053911\\83	0.0053704\\84	0.0053497\\85	0.0053296\\86	0.0053171\\87	0.0053089\\88	0.0053020\\89	0.0052682\\90	0.0052582\\91	0.0052370\\92	0.0052217\\93	0.0052266\\94	0.0052146\\95	0.0052278\\96	0.0052153\\97	0.0052009\\98	0.0051880\\99	0.0051778\\100	0.0051623\\101	0.0051794\\102	0.0051963\\103	0.0051662\\104	0.0051817\\105	0.0051702\\106	0.0051359\\107	0.0051724\\108	0.0051643\\109	0.0051626\\110	0.0051793\\111	0.0051704\\112	0.0051594\\113	0.0051558\\114	0.0051609\\115	0.0051503\\116	0.0051637\\117	0.0051838\\118	0.0051820\\119	0.0051747\\120	0.0051988\\121	0.0051781\\122	0.0051610\\123	0.0052025\\124	0.0051883\\125	0.0051843\\126	0.0052103\\127	0.0052031\\128	0.0052213\\129	0.0052140\\130	0.0052129\\131	0.0052183\\132	0.0052126\\133	0.0052207\\134	0.0051998\\135	0.0052205\\136	0.0052569\\137	0.0052496\\138	0.0052447\\139	0.0052644\\140	0.0052824\\141	0.0052493\\142	0.0052534\\143	0.0047938\\144	0.0025658\\145	0.0004532\\146	0.0001242\\147	0.0001408\\148	0.0001160\\149	0.0001044\\150	0.0000889\\151	0.0001246\\152	0.0001139\\153	0.0001166\\154	0.0000993\\155	0.0000932\\156	0.0000878\\157	0.0000783\\158	0.0000859\\159	0.0000766\\160	0.0000719\\161	0.0000758\\162	0.0000649\\163	0.0000609\\164	0.0000631\\165	0.0000570\\166	0.0000467\\167	0.0000496\\168	0.0000455\\169	0.0000446\\170	0.0000359\\171	0.0000350\\172	0.0000358\\173	0.0000340\\174	0.0000263\\175	0.0000203\\176	0.0000188\\177	0.0000169\\178	0.0000150\\179	0.0000130\\180	0.0000069\\181	0.0000046\\182	0.0000037\\183	0.0000021\\184	-0.0000033\\185	-0.0000051\\186	-0.0000099\\187	-0.0000067\\188	-0.0000127\\189	-0.0000190\\190	-0.0000180\\191	-0.0000189\\192	-0.0000262\\193	-0.0000341\\194	-0.0000288\\195	-0.0000259\\196	-0.0000321\\197	-0.0000254\\198	-0.0000253\\199	-0.0000353\\200	-0.0000314\\201	-0.0000394\\202	-0.0000472\\203	-0.0000469\\204	-0.0000434\\205	-0.0000420\\206	-0.0000519\\207	-0.0000561\\208	-0.0000529\\209	-0.0000493\\210	-0.0000497\\211	-0.0000553\\212	-0.0000583\\213	-0.0000512\\214	-0.0000584\\215	-0.0000551\\216	-0.0000661\\217	-0.0000734\\218	-0.0000788\\219	-0.0000633\\220	-0.0000235\\221	-0.0000519\\222	-0.0000424\\223	-0.0000523\\224	-0.0000390\\225	-0.0000560\\226	0.0005071\\227	0.0028020\\228	0.0050925\\229	0.0055388\\230	0.0054900\\231	0.0054981\\232	0.0055181\\233	0.0055060\\234	0.0055185\\235	0.0055336\\236	0.0055545\\237	0.0055182\\238	0.0055430\\239	0.0055498\\240	0.0055318\\241	0.0055458\\242	0.0055286\\243	0.0055053\\244	0.0054980\\245	0.0055273\\246	0.0055372\\247	0.0055043\\248	0.0055445\\249	0.0055585\\250	0.0055189\\251	0.0055295\\252	0.0055210\\253	0.0054931\\254	0.0054963\\255	0.0055177\\256	0.0055101\\257	0.0055306\\258	0.0055002\\259	0.0054979\\260	0.0055207\\261	0.0054559\\262	0.0054857\\263	0.0054970\\264	0.0054589\\265	0.0055058\\266	0.0054847\\267	0.0054652\\268	0.0054494\\269	0.0054376\\270	0.0054094\\271	0.0053938\\272	0.0054511\\273	0.0054303\\274	0.0054124\\275	0.0053972\\276	0.0053672\\277	0.0053456\\278	0.0053422\\279	0.0053180\\280	0.0052981\\281	0.0053311\\282	0.0052813\\283	0.0052493\\284	0.0052422\\285	0.0051723\\286	0.0051720\\287	0.0051693\\288	0.0051282\\289	0.0050768\\290	0.0050466\\291	0.0049856\\292	0.0049769\\293	0.0049121\\294	0.0048270\\295	0.0048266\\296	0.0047195\\297	0.0046281\\298	0.0045813\\299	0.0044590\\300	0.0043383\\301	0.0042379\\302	0.0040863\\303	0.0038491\\304	0.0036442\\305	0.0033536\\306	0.0028973\\307	0.0023412\\308	0.0013232\\309	-0.0027149\\310	-0.0057936\\311	-0.0060375\\312	-0.0044033\\313	-0.0036578\\314	-0.0031685\\315	-0.0028000\\316	-0.0026068\\317	-0.0023850\\318	-0.0022009\\319	-0.0020927\\320	-0.0019472\\321	-0.0018354\\322	-0.0017811\\323	-0.0016910\\324	-0.0016201\\325	-0.0015417\\326	-0.0014846\\327	-0.0014185\\328	-0.0013617\\329	-0.0013535\\330	-0.0012958\\331	-0.0012573\\332	-0.0012127\\333	-0.0011803\\334	-0.0011420\\335	-0.0011118\\336	-0.0011061\\337	-0.0010726\\338	-0.0010503\\339	-0.0010152\\340	-0.0009853\\341	-0.0009668\\342	-0.0009430\\343	-0.0009438\\344	-0.0009298\\345	-0.0009175\\346	-0.0008887\\347	-0.0008570\\348	-0.0008458\\349	-0.0008107\\350	-0.0008231\\351	-0.0008165\\352	-0.0007950\\353	-0.0007925\\354	-0.0007717\\355	-0.0007659\\356	-0.0007603\\357	-0.0007485\\358	-0.0007386\\359	-0.0007123\\360	-0.0006893\\361	-0.0006875\\362	-0.0006699\\363	-0.0006640\\364	-0.0006804\\365	-0.0006719\\366	-0.0006498\\367	-0.0006485\\368	-0.0006412\\369	-0.0006321\\370	-0.0006444\\371	-0.0006284\\372	-0.0006193\\373	-0.0005667\\
	};
\addplot [color=cyan, line width=1.0pt, dashed] table[row sep=crcr]{%
0	0.000000\\1	0.000422\\2	0.000292\\3	0.000077\\4	0.000099\\5	0.000160\\6	0.000047\\7	0.000011\\8	0.000179\\9	0.000055\\10	0.000007\\11	0.000208\\12	0.000082\\13	0.000000\\14	0.000000\\15	0.000000\\16	0.000000\\17	0.000000\\18	0.000000\\19	0.000000\\20	0.000000\\21	0.000000\\22	0.000000\\23	0.000000\\24	0.000000\\25	0.000000\\26	0.000000\\27	0.000000\\28	0.000000\\29	0.000000\\30	0.000000\\31	0.000000\\32	0.000000\\33	0.000000\\34	0.000000\\35	0.000000\\36	0.000000\\37	0.000000\\38	0.000000\\39	0.000000\\40	0.000000\\41	0.000000\\42	0.000000\\43	0.000000\\44	0.000000\\45	0.000000\\46	0.000000\\47	0.000000\\48	0.000000\\49	0.000000\\50	0.000000\\51	0.000000\\52	0.000000\\53	0.000000\\54	0.000000\\55	0.000000\\56	0.000000\\57	0.000000\\58	0.002121\\59	0.035558\\60	0.108415\\61	0.234690\\62	0.511082\\63	0.717749\\64	0.849809\\65	0.895947\\66	0.918909\\67	0.935885\\68	0.948179\\69	0.956851\\70	0.963757\\71	0.969214\\72	0.973356\\73	0.976005\\74	0.977977\\75	0.980320\\76	0.981882\\77	0.983213\\78	0.984346\\79	0.986401\\80	0.986514\\81	0.986924\\82	0.988354\\83	0.987803\\84	0.988292\\85	0.988921\\86	0.988623\\87	0.990099\\88	0.990684\\89	0.990430\\90	0.990719\\91	0.990117\\92	0.989388\\93	0.991070\\94	0.991127\\95	0.991773\\96	0.993796\\97	0.991710\\98	0.992123\\99	0.991699\\100	0.990886\\101	0.991904\\102	0.992841\\103	0.992548\\104	0.992988\\105	0.993125\\106	0.991553\\107	0.992292\\108	0.992406\\109	0.991800\\110	0.992569\\111	0.992526\\112	0.992197\\113	0.991809\\114	0.992183\\115	0.991691\\116	0.990832\\117	0.992360\\118	0.991545\\119	0.990644\\120	0.992573\\121	0.991423\\122	0.990632\\123	0.990098\\124	0.990332\\125	0.990615\\126	0.990104\\127	0.991260\\128	0.991027\\129	0.990664\\130	0.991866\\131	0.990750\\132	0.991054\\133	0.990962\\134	0.988437\\135	0.988505\\136	0.987749\\137	0.985933\\138	0.980749\\139	0.972786\\140	0.962498\\141	0.939735\\142	0.897450\\143	0.807191\\144	0.620390\\145	0.319166\\146	0.148814\\147	0.077775\\148	0.045319\\149	0.027831\\150	0.016292\\151	0.012070\\152	0.008017\\153	0.005862\\154	0.004215\\155	0.002561\\156	0.002545\\157	0.003043\\158	0.003407\\159	0.003529\\160	0.003620\\161	0.003613\\162	0.003545\\163	0.003498\\164	0.003295\\165	0.003384\\166	0.003139\\167	0.003174\\168	0.002959\\169	0.002825\\170	0.002804\\171	0.002983\\172	0.002814\\173	0.002832\\174	0.002715\\175	0.002631\\176	0.002590\\177	0.002625\\178	0.002638\\179	0.002668\\180	0.002669\\181	0.002601\\182	0.002790\\183	0.002826\\184	0.002825\\185	0.002755\\186	0.002800\\187	0.002767\\188	0.002772\\189	0.002585\\190	0.002558\\191	0.002620\\192	0.002494\\193	0.002376\\194	0.002343\\195	0.002360\\196	0.002356\\197	0.002346\\198	0.002160\\199	0.002374\\200	0.002251\\201	0.002212\\202	0.002323\\203	0.002230\\204	0.002252\\205	0.002208\\206	0.002205\\207	0.002058\\208	0.001958\\209	0.001909\\210	0.001691\\211	0.001438\\212	0.001180\\213	0.000872\\214	0.001885\\215	0.003781\\216	0.005460\\217	0.008302\\218	0.012160\\219	0.015854\\220	0.025687\\221	0.035833\\222	0.051751\\223	0.072042\\224	0.105961\\225	0.158081\\226	0.239758\\227	0.442472\\228	0.650456\\229	0.791312\\230	0.853377\\231	0.893739\\232	0.919913\\233	0.938902\\234	0.951221\\235	0.961150\\236	0.970056\\237	0.973461\\238	0.977564\\239	0.981458\\240	0.982924\\241	0.985244\\242	0.986034\\243	0.985956\\244	0.986423\\245	0.987362\\246	0.988330\\247	0.987176\\248	0.987694\\249	0.988815\\250	0.987315\\251	0.986687\\252	0.987033\\253	0.986008\\254	0.985737\\255	0.985482\\256	0.986221\\257	0.985768\\258	0.984350\\259	0.984818\\260	0.984016\\261	0.983199\\262	0.983785\\263	0.983726\\264	0.983646\\265	0.983681\\266	0.983979\\267	0.983541\\268	0.982208\\269	0.982363\\270	0.981213\\271	0.980844\\272	0.982217\\273	0.982625\\274	0.981113\\275	0.981721\\276	0.980612\\277	0.978503\\278	0.980128\\279	0.979313\\280	0.979011\\281	0.979448\\282	0.978181\\283	0.977268\\284	0.977207\\285	0.975542\\286	0.974552\\287	0.974343\\288	0.972634\\289	0.970995\\290	0.969245\\291	0.966477\\292	0.965410\\293	0.962229\\294	0.958213\\295	0.954992\\296	0.949687\\297	0.943674\\298	0.936673\\299	0.928084\\300	0.917069\\301	0.903371\\302	0.885292\\303	0.861747\\304	0.832475\\305	0.793858\\306	0.732458\\307	0.642620\\308	0.515539\\309	0.209843\\310	0.026234\\311	0.000000\\312	0.000000\\313	0.000000\\314	0.000000\\315	0.000000\\316	0.000000\\317	0.000000\\318	0.000000\\319	0.000000\\320	0.000000\\321	0.000000\\322	0.000000\\323	0.000000\\324	0.000000\\325	0.000000\\326	0.000000\\327	0.000000\\328	0.000000\\329	0.000000\\330	0.000000\\331	0.000000\\332	0.000000\\333	0.000000\\334	0.000000\\335	0.000000\\336	0.000000\\337	0.000000\\338	0.000000\\339	0.000000\\340	0.000000\\341	0.000000\\342	0.000000\\343	0.000000\\344	0.000000\\345	0.000000\\346	0.000000\\347	0.000000\\348	0.000000\\349	0.000000\\350	0.000000\\351	0.000000\\352	0.000000\\353	0.000000\\354	0.000000\\355	0.000000\\356	0.000000\\357	0.000000\\358	0.000000\\359	0.000000\\360	0.000000\\361	0.000000\\362	0.000000\\363	0.000000\\364	0.000000\\365	0.000000\\366	0.000000\\367	0.000000\\368	0.000000\\369	0.000000\\370	0.000000\\371	0.000000\\372	0.000000\\373	0.000000\\
};
\addplot [color=teal, line width=1.0pt, dashed] table[row sep=crcr]{%
0	0.000362\\1	0.000357\\2	0.000347\\3	0.000340\\4	0.000323\\5	0.000306\\6	0.000289\\7	0.000267\\8	0.000242\\9	0.000216\\10	0.000190\\11	0.000166\\12	0.000141\\13	0.000113\\14	0.000093\\15	0.000072\\16	0.000053\\17	0.000038\\18	0.000023\\19	0.000015\\20	0.000004\\21	5.513E-7\\22	0.000003\\23	-0.000004\\24	0.000012\\25	0.000006\\26	0.000007\\27	0.000005\\28	0.000001\\29	0.000000\\30	0.000000\\31	0.000000\\32	0.000000\\33	0.000000\\34	0.000000\\35	0.000000\\36	0.000000\\37	0.000000\\38	0.000000\\39	0.000000\\40	0.000000\\41	0.000000\\42	0.000000\\43	0.000000\\44	0.000000\\45	0.000000\\46	0.000000\\47	0.000000\\48	0.000000\\49	0.000005\\50	0.000002\\51	-0.000006\\52	0.000005\\53	0.000006\\54	0.000005\\55	0.000019\\56	0.000015\\57	0.000026\\58	0.002238\\59	0.036072\\60	0.106199\\61	0.231496\\62	0.508071\\63	0.717107\\64	0.850208\\65	0.896483\\66	0.918893\\67	0.935849\\68	0.948063\\69	0.956665\\70	0.963601\\71	0.969077\\72	0.972827\\73	0.975940\\74	0.978327\\75	0.980099\\76	0.982159\\77	0.983539\\78	0.984631\\79	0.985980\\80	0.986519\\81	0.987170\\82	0.988014\\83	0.988323\\84	0.988807\\85	0.989258\\86	0.989553\\87	0.989884\\88	0.990158\\89	0.990369\\90	0.990572\\91	0.990760\\92	0.990968\\93	0.991248\\94	0.991477\\95	0.991682\\96	0.991860\\97	0.991980\\98	0.992101\\99	0.992213\\100	0.992287\\101	0.992342\\102	0.992380\\103	0.992403\\104	0.992414\\105	0.992424\\106	0.992427\\107	0.992431\\108	0.992426\\109	0.992421\\110	0.992405\\111	0.992378\\112	0.992340\\113	0.992280\\114	0.992198\\115	0.992107\\116	0.992011\\117	0.991917\\118	0.991814\\119	0.991710\\120	0.991612\\121	0.991497\\122	0.991362\\123	0.991222\\124	0.991113\\125	0.991002\\126	0.990892\\127	0.990794\\128	0.990703\\129	0.990626\\130	0.990571\\131	0.990536\\132	0.990463\\133	0.990279\\134	0.989699\\135	0.988951\\136	0.987892\\137	0.985975\\138	0.981076\\139	0.973433\\140	0.962418\\141	0.940300\\142	0.898218\\143	0.807425\\144	0.618931\\145	0.318686\\146	0.147692\\147	0.076399\\148	0.044836\\149	0.027477\\150	0.016455\\151	0.011415\\152	0.007539\\153	0.005132\\154	0.003762\\155	0.003730\\156	0.003712\\157	0.003696\\158	0.003675\\159	0.003647\\160	0.003609\\161	0.003555\\162	0.003501\\163	0.003445\\164	0.003383\\165	0.003324\\166	0.003271\\167	0.003215\\168	0.003132\\169	0.003086\\170	0.003052\\171	0.003015\\172	0.002962\\173	0.002932\\174	0.002908\\175	0.002892\\176	0.002885\\177	0.002881\\178	0.002879\\179	0.002872\\180	0.002871\\181	0.002869\\182	0.002871\\183	0.002872\\184	0.002864\\185	0.002866\\186	0.002864\\187	0.002853\\188	0.002848\\189	0.002833\\190	0.002815\\191	0.002786\\192	0.002743\\193	0.002708\\194	0.002683\\195	0.002654\\196	0.002646\\197	0.002629\\198	0.002610\\199	0.002583\\200	0.002556\\201	0.002531\\202	0.002506\\203	0.002474\\204	0.002432\\205	0.002372\\206	0.002299\\207	0.002239\\208	0.002200\\209	0.002184\\210	0.002164\\211	0.002156\\212	0.002165\\213	0.002157\\214	0.002358\\215	0.003397\\216	0.005578\\217	0.008376\\218	0.012189\\219	0.016875\\220	0.025007\\221	0.035328\\222	0.050505\\223	0.071251\\224	0.104987\\225	0.157075\\226	0.240319\\227	0.442434\\228	0.650869\\229	0.792664\\230	0.854933\\231	0.894752\\232	0.920811\\233	0.939356\\234	0.951858\\235	0.961715\\236	0.969731\\237	0.973953\\238	0.977876\\239	0.981158\\240	0.983334\\241	0.985107\\242	0.986069\\243	0.986679\\244	0.987161\\245	0.987361\\246	0.987396\\247	0.987395\\248	0.987394\\249	0.987372\\250	0.987317\\251	0.987201\\252	0.987044\\253	0.986827\\254	0.986533\\255	0.986248\\256	0.985961\\257	0.985600\\258	0.985278\\259	0.985028\\260	0.984758\\261	0.984550\\262	0.984399\\263	0.984248\\264	0.984136\\265	0.984030\\266	0.983879\\267	0.983676\\268	0.983455\\269	0.983214\\270	0.982922\\271	0.982618\\272	0.982396\\273	0.982181\\274	0.981904\\275	0.981687\\276	0.981374\\277	0.980969\\278	0.980611\\279	0.980115\\280	0.979573\\281	0.979165\\282	0.978636\\283	0.978005\\284	0.977284\\285	0.976258\\286	0.975186\\287	0.974083\\288	0.972852\\289	0.971349\\290	0.969633\\291	0.967548\\292	0.965451\\293	0.962348\\294	0.958601\\295	0.954958\\296	0.949981\\297	0.944093\\298	0.937084\\299	0.928538\\300	0.917601\\301	0.903922\\302	0.885988\\303	0.862659\\304	0.833579\\305	0.794997\\306	0.733834\\307	0.644069\\308	0.516611\\309	0.211103\\310	0.027463\\311	0.000009\\312	0.000003\\313	0.000003\\314	0.000004\\315	0.000013\\316	0.000003\\317	0.000004\\318	0.000000\\319	0.000000\\320	0.000000\\321	0.000000\\322	0.000000\\323	0.000000\\324	0.000000\\325	0.000000\\326	0.000000\\327	0.000000\\328	0.000000\\329	0.000000\\330	0.000000\\331	0.000000\\332	0.000000\\333	0.000000\\334	0.000000\\335	0.000000\\336	0.000000\\337	0.000000\\338	0.000000\\339	0.000000\\340	0.000000\\341	0.000000\\342	0.000000\\343	0.000000\\344	0.000000\\345	0.000000\\346	0.000000\\347	0.000000\\348	0.000000\\349	0.000000\\350	0.000000\\351	0.000000\\352	0.000000\\353	0.000000\\354	0.000000\\355	0.000000\\356	0.000000\\357	0.000000\\358	0.000000\\359	0.000000\\360	0.000000\\361	0.000000\\362	0.000000\\363	0.000000\\364	0.000000\\365	0.000000\\366	0.000000\\367	0.000000\\368	0.000000\\369	0.000000\\370	0.000000\\371	0.000000\\372	0.000000\\373	0.000000\\
};
%
\end{axis} 
 \end{tikzpicture}